\documentclass{INTERSPEECH2023}

\usepackage{latexsym}
\usepackage{url}
\usepackage{amsthm}
\usepackage{amssymb}
\usepackage{graphicx}
\usepackage{booktabs}
\usepackage{placeins}
\usepackage{multirow}
\usepackage{tabularx}
\usepackage{epstopdf}
\usepackage{color}
\usepackage{subcaption}
\usepackage{amsfonts}
\usepackage{amsmath}
\usepackage{balance}
\usepackage{enumitem}
\usepackage{xcolor}
\usepackage{wrapfig}
\usepackage{graphics}
\usepackage{todonotes}
\usepackage{makecell}
\usepackage{bbm}
\usepackage{bm}

\interspeechcameraready

\title{Label Aware Speech Representation Learning For Language Identification}
\name{Shikhar Vashishth, Shikhar Bharadwaj, Sriram Ganapathy, Ankur Bapna, \\ Min Ma, Wei Han, Vera Axelrod, Partha Talukdar}
\address{
  Google Research}
\email{ \{shikharv,shikharop,srigana,ankurbpn,minm,weihan,vaxelrod,partha\}@google.com}

\begin{document}
\maketitle
\begin{abstract}
Speech representation learning approaches for non-semantic tasks such as language recognition have either explored supervised embedding extraction methods using a classifier model or self-supervised representation learning approaches  using raw data. In this paper, we propose a novel framework of combining self-supervised representation learning with the language label information for the pre-training task. This framework, termed as \textbf{L}abel \textbf{A}ware \textbf{S}peech \textbf{R}epresentation  (LASR) learning, uses a triplet based objective function to incorporate language labels along with the self-supervised loss function. The speech representations are further fine-tuned for the downstream task.  The language recognition experiments are performed on two public datasets --  FLEURS and Dhwani. In these experiments, we illustrate that the proposed LASR framework improves over the state-of-the-art systems on language identification. We also report an analysis of the robustness of LASR approach to noisy/missing labels as well as its application to  multi-lingual speech recognition tasks.
\end{abstract}

\newcommand{\refalg}[1]{Algorithm \ref{#1}}
\newcommand{\refeqn}[1]{Equation \ref{#1}}
\newcommand{\reffig}[1]{Figure \ref{#1}}
\newcommand{\reftbl}[1]{Table \ref{#1}}
\newcommand{\refsec}[1]{Section \ref{#1}}

\newcommand{\reminder}[1]{\textcolor{red}{[[ #1 ]]}\typeout{#1}}
\newcommand{\reminderR}[1]{\textcolor{gray}{[[ #1 ]]}\typeout{#1}}

\newcommand{\add}[1]{\textcolor{red}{#1}\typeout{#1}}
\newcommand{\remove}[1]{\sout{#1}\typeout{#1}}

\newcommand{\method}{LASR}

\newcommand{\problem}{DD}
\newcommand{\problemfull}{Document Dating}

\newcommand{\mc}[1]{\mathcal{#1}}
\newcommand{\bmm}[1]{\bm{\mathcal{#1}}}
\newcommand{\real}[1]{\mathbb{R}^{#1}}

\newcommand{\tensor}{\mathcal{X}}
\newcommand{\Real}{\mathbb{R}}

\newcommand{\tuples}{\mathbb{T}}

\newcommand{\argmax}{arg\,max}

\newcommand\norm[1]{\left\lVert#1\right\rVert}

\newcommand{\note}[1]{\textcolor{blue}{#1}}

\newcommand*{\Scale}[2][4]{\scalebox{#1}{$#2$}}%
\newcommand*{\Resize}[2]{\resizebox{#1}{!}{$#2$}}%
\definecolor{officegreen}{rgb}{0.0, 0.5, 0.0}
\def\mat#1{\mbox{\bf #1}}
\section{Introduction}
\label{sec:introduction}
The conventional approach for deriving speech representations for non-semantic speech tasks, such as speaker and language recognition, involved the use of training deep neural models with a statistics pooling layer. 
Some of the popular methods in this direction include  d-vectors \cite{heigold2016end} and x-vectors \cite{snyder2018x}, where a deep neural model is trained to classify the speaker/language labels on a large corpus of supervised data.   
However, recent trends in speech processing has seen a paradigm shift towards self-supervision based representation learning, mirroring the efforts in computer vision \cite{vision_unsup1} and natural language processing \cite{bert}. Some popular examples of such approaches include contrastive predictive coding (CPC) \cite{oord2018representation}, wav2vec family of models \cite{wave2vec2,w2v_bert}, and hidden unit BERT (HuBERT) \cite{hubert}.  These methods primarily rely on learning  speech representations at the frame-level with its impact reported on semantic tasks such as low-resource speech recognition   \cite{hubert,baevski2021unsupervised} or zero resource spoken language modeling  \cite{maekaku2022exploration}. These representations have also been investigated for speaker and language recognition tasks \cite{fan2020exploring} through various benchmarks such as SUPERB \cite{yang2021superb} and NOSS \cite{trill}. 

In many learning paradigms, it is plausible to have portions of pre-training data along with the corresponding meta-data. 
In the broad spectrum of representation learning, where supervised and self-supervised frameworks constitute the two-ends of the spectrum, we hypothesize that a combination of supervision and self-supervision based methods may be more optimal than either of the two frameworks in isolation, for scenarios where parts of the pre-training have additional meta-data in the form of labels. In this paper, we propose a framework for \textbf{L}abel \textbf{A}ware \textbf{S}peech \textbf{R}epresentation learning (LASR) for such scenarios. To the best of our knowledge, this is the first attempt to combine label information with a self-supervision loss for non-semantic speech tasks. The contributions from this work are as follows.
\begin{enumerate}[itemsep=2pt,parsep=2pt,partopsep=2pt,leftmargin=*,topsep=2pt]
	\item We propose \method{}, a framework for incorporating label information in self-supervised speech representation learning.
	\item We demonstrate the effectiveness of \method{} for  language identification task and establish its efficacy even with missing and noisy labels. 
	\item Our findings demonstrate that inclusion of language information in the pre-training phase results in state-of-art-results on the FLEURS dataset \cite{dataset_fleurs}.
\end{enumerate}
\section{Related Work}
\label{sec:related_works}


\begin{figure*}[t]
	\centering
	\includegraphics[scale=0.8]{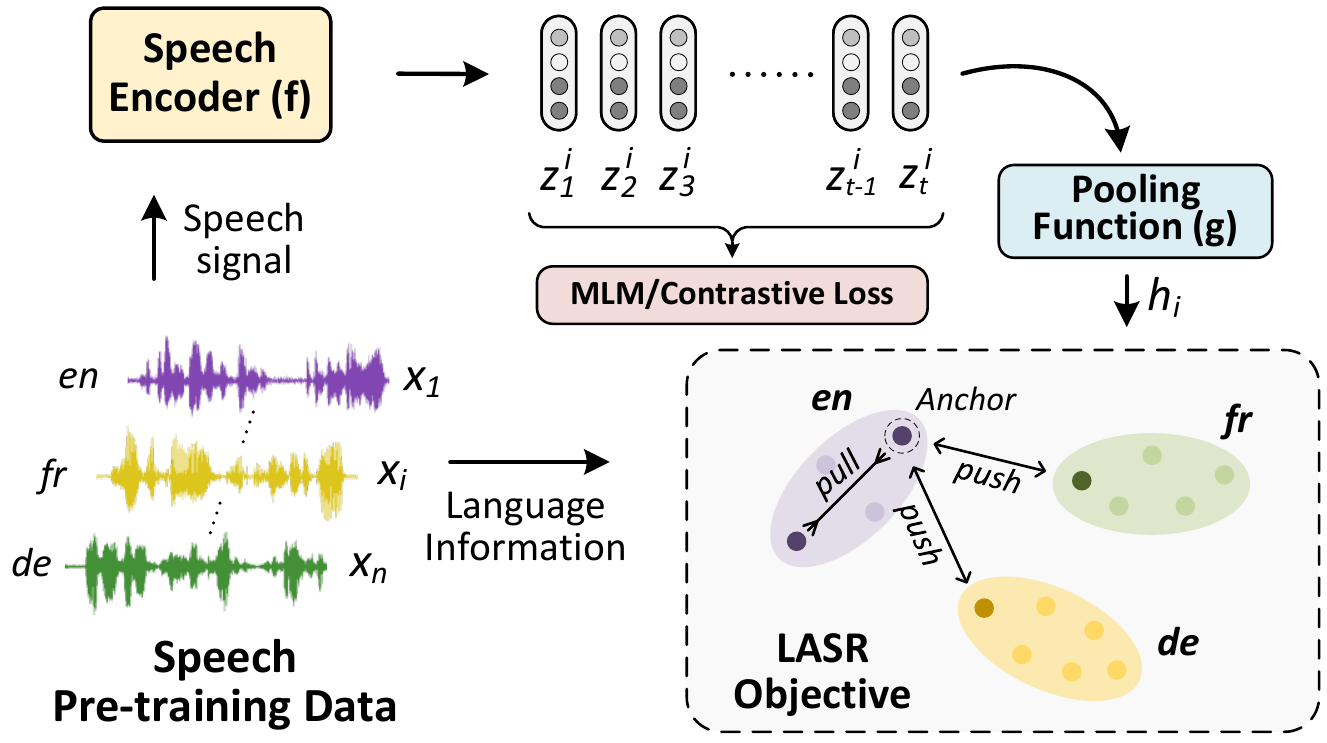}
	\caption{\label{fig:overview}\small Overview of \method{} framework. Given a batch of multilingual speech samples, for each sample $x_i$, \method{} utilizes a speech encoder ($f$) to obtain frame-level representations ${z^i_1, z^i_2, ..., z^i_t}$. These are used for computing self-supervised loss and are fed to a pooling function ($g$) to derive utterance-level embedding $h_i$. The language labels of samples and $h_i$'s are used to compute \method{} loss.}
	\vspace{-0.15in}
\end{figure*}

\textbf{Supervised Learning:} Deep learning methods for non-semantic speech tasks initially explored speech recognition models in the unsupervised i-vector framework  \cite{lei2014novel}. Further, the embeddings derived from a classifier model, trained on large amounts of supervised pre-training data, showed promising results for speaker \cite{heigold2016end} and language recognition \cite{snyder2018spoken}. The initial architecture based on time-delay neural network (TDNN)  \cite{snyder2018x} has since been improved with factorization \cite{povey2018semi}, residual networks \cite{li2017deep} and more recently with channel attention based TDNN \cite{desplanques2020ecapa}. 
Most of these approaches use a pooling layer to convert frame level representations to an utterance level embedding followed by a cross-entropy based classification objective.
However, our work investigates the combination of self-supervision objectives along with the supervised labels. \\

\noindent \textbf{Speech Self-Supervised Learning:} Prior research in the field of speech self-supervised learning can largely be classified into two major categories: contrastive and predictive. The contrastive approaches learn by maximizing the similarity of an anchor with the positive samples, while simultaneously minimizing its similarity with the negative samples. The class of wav2vec models \cite{wave2vec2,wave2vec}   fall in this category. On the other hand, predictive methods are based on masked language modeling (MLM) objective  \cite{bert}. The examples  include Discrete-BERT \cite{discrete_bert}, w2v-BERT \cite{w2v_bert}, HuBERT \cite{hubert}, and BEST-RQ \cite{best_rq}. Our proposed framework enables integration of label  information in both categories of methods.\\

\noindent \textbf{Non-Semantic Speech Representations:} For tasks such as language identification, speaker diarization, and emotion detection, it is essential to also capture the non-semantic aspect of speech. TRILL \cite{trill} utilizes temporal proximity as supervision signal to learn non-semantic representation, with promising results on   NOSS (non-semantic speech) benchmark. Further, methods such as FRILL \cite{frill} and TRILLsson \cite{trillson} have  enhanced the  performance and efficiency of these models. Another approach named COLA \cite{cola} modifies the negative sampling scheme to learn more general purpose audio representation. 
All these works are specifically designed for contrastive techniques, whereas \method{} can be integrated with any self-supervised speech representation learning method. \\

\noindent \textbf{Joint learning:} Talnikar et. al. \cite{talnikar2021joint} explored the combination of supervised (connectionist temporal cost (CTC)) and self-supervised (contrastive prediction loss (CPC)) losses for speech recognition. Similarly, UniSpeech \cite{wang2021unispeech} used CTC labeling and phonetically-aware contrastive learning in a multi-task learning framework. Bai et. al. 
\cite{bai2022joint} used the self-supervised MLM loss and the speech recognition loss for a multi-lingual speech recognition system. However, all these approaches learn frame-level representations for a semantic task. In our work, the LASR framework combines utterance-level label supervision with frame-level self-supervision. 

\section{LASR Framework}
\label{sec:details}
A comprehensive illustration of the \method{} framework is depicted in Figure \ref{fig:overview}.
A self-supervised speech encoding model $f:\bm{x} \rightarrow \bmm{Z}$, such as wave2vec-2.0 \cite{wave2vec2} or w2v-BERT \cite{w2v_bert}, transforms a raw audio waveform $\bm{\mathcal{X}}$ into the frame-level speech representations $\bmm{Z} = [\bm{z}_1, \bm{z}_2, ..., \bm{z}_T]$. 
In our proposed LASR framework, the pre-training  dataset is denoted as $\mc{D}=\{(\bmm{X}_1, l_1), (\bmm{X}_2, l_2),..., (\bmm{X}_n, l_n)\}$, where each speech utterance $\bmm{X}_i$ is accompanied by its corresponding language label $l_i$. The remaining unlabeled samples will solely be utilized for optimizing the self-supervised objective.


\begin{table*}[t!]
	\centering
	\begin{tabular}{lcccccccccc}
	    \toprule
	     & \multicolumn{5}{c}{\textbf{FLEURS}} & \multicolumn{5}{c}{\textbf{Dhwani}}  \\
	    \cmidrule(r){2-6} \cmidrule(r){7-11}
	    \multicolumn{1}{l}{\textbf{Method}} & O (48) & NO (54) & \multicolumn{3}{c}{Overall}  & O (5) & NO (17) & \multicolumn{3}{c}{Overall} \\
	    \cmidrule(r){2-2} \cmidrule(r){3-3} \cmidrule(r){4-6} \cmidrule(r){7-7} \cmidrule(r){8-8} \cmidrule(r){9-11}
	     & Acc & Acc & Acc & F1 & EER & Acc  & Acc & Acc & F1 & EER \\
		\midrule
		
		wav2vec 2.0 \cite{wave2vec2} &  84.8 & 72.2 & 78.2 & 76.3 & 1.1 & 77.6 & 47.6 & 56.0 & 41.1 & 15.9 \\
		w2v-BERT \cite{w2v_bert} & 87.7 & 69.6 & 78.0 & 77.7 & 0.5 & 78.8 & 49.9 & 58.0 & 42.6 & 15.4 \\
        BEST-RQ \cite{best_rq} &  86.8 & 65.6 & 75.8 & 73.3 & 1.2 & 76.2 & 46.4 & 54.7 & 39.8 & 16.9 \\
        \midrule
        \method{} + wav2vec 2.0 & 88.4 & \textbf{76.5} & \textbf{82.1} & \textbf{80.4} & 0.7 & \textbf{80.7} & 50.1 & 58.7 & 43.3 & 16.3 \\
        \method{} + w2v-BERT & 88.9 & 74.3 & 81.3 & \textbf{80.4} & \textbf{0.5} & 78.1 & \textbf{52.2} & \textbf{59.5} & \textbf{44.2} & \textbf{15.2} \\
		\method{} + BEST-RQ & \textbf{90.6} & 73.4 & 81.6 & 79.7 & \textbf{0.5} & 77.0 & 50.8 & 58.2 & 43.2 & 16.1 \\
		\bottomrule
	\end{tabular}
	\caption{\label{tbl:langid_main} Language identification accuracy (\%), macro-F1 and equal error rate (EER) for various approaches. \textit{O} stands for languages that overlap with the pre-training data and \textit{NO} are the non-overlapping languages. In parenthesis, we report the number of classes in each category. We find that methods trained with the LASR objective achieve better performance. Refer to Section \ref{sec:results} for details.}
	\vspace{-0.2in}
\end{table*}

Subsequently, we employ an aggregation function $g: \bmm{Z} \rightarrow \bm{h}$ to obtain an utterance level embedding $\bm{h} = g(\bmm{Z})$. Here, $g$ can, in general, take the form of a neural network such as LSTM or an attention model \cite{attention}. In our case, $g$ is an average pooling, i.e.,
$\bm{h} = g(\bmm{Z}) = \dfrac{1}{T} \sum_{t}{\bm{z}_t}.$

For an anchor speech utterance $\bmm{X}_i$ with aggregate representation $\bm{h}_i$ and language label $l_i$, we select a positive and negative sample: $\bmm{X}^+_i$ and $\bmm{X}^-_i$ such that $l^+_i = l_i$ and $l^-_i \neq l_i$.  We use the triplet-loss objective, as proposed in \cite{loss_semi_hard_triplet}, i.e.,
\begin{equation}\label{eq:triplet}
\mc{L}_{\texttt{tri}}  = \sum_{i}{\max \big[0,  \gamma + d(\bm{h}_i, \bm{h}_i^+) - d(\bm{h}_i, \bm{h}_i^-) \big]},
\end{equation} 
where $\gamma$ is the margin and $d(\cdot, \cdot)$ is the distance metric employed. In this work, we use angular distance as the distance metric. We also explore the hard triplet mining strategy \cite{loss_hard_triplet,triplet_loss_lang}, where the most distant positive and closest negative sample within the mini-batch are selected to form the triplet. 
\begin{equation}\label{eq:hard}
\mc{L}_{\texttt{hard}} = \sum_{i} \max [0, \gamma + \max_{j \in i^+}  d(\bm{h}_i, \bm{h}_j) - \min_{j \in i^-} d (\bm{h}_i, \bm{h}_j) ]
\end{equation} 
Here, ${j \in i^+}$ denotes the set of utterances  in the mini-batch that have the same label $l_j= l_i$ and ${j \in i^-}$ denotes the set of utterances with a different label, i.e., $l_j \neq l_i$. The total loss function used in the proposed \method{} approach is given by, 
\begin{equation} \label{eq:hard-triplet}
\mc{L} _{\texttt{\method{}}} = \mc{L}_{\texttt{SSL}} + \lambda \cdot \mc{L}_{\texttt{hard}}.
\end{equation}
Here, $\mc{L}_{\texttt{SSL}}$ is the loss corresponding to the self-supervised speech encoding method $f$ and $\lambda$ decides the trade-off between the SSL objective and hard-triplet objective. In our experiments, we find that having the SSL objective is crucial for achieving the best language recognition performance. In Section \ref{sec:discussion}, we also assess the significance of altering the parameter $\lambda$. In addition to the triplet loss (Eq. \ref{eq:hard-triplet}), we also examine generalized end-to-end loss \cite{loss_ge2e}. 
\begin{equation}\label{eq:ge2e}
\mc{L}_{\texttt{ge2e}} = \sum_{i} 1  - \sigma (\max_{j \in i^+}  d(\bm{h}_i, \bm{h}_j) ) +  \sigma  ( \min_{j \in i^-} d (\bm{h}_i, \bm{h}_j)  ).
\end{equation} 
\section{Experimental Setup}
\label{sec:experiments}

\subsection{Dataset}
\label{sec:datasets}

\textbf{Pre-training Data:}
In our experiments, we employ a large set of open source speech data for pre-training, totaling about $429$k audio hours. This consists of $372$k hours of speech data across $23$ languages from VoxPopuli dataset \cite{dataset_vp}, $50$k hours of speech from $25$ languages in Common Voice dataset \cite{dataset_cv}, $50$k hours of read speech in $8$ European languages from Multilingual LibriSpeech (MLS) corpus \cite{dataset_mls}, and $1000$ hours of telephone conversation data across $17$ African and Asian languages from BABEL dataset \cite{dataset_babel}. Overall, this combined dataset has speech utterances from 75 languages.

\noindent \textbf{Evaluation Data:}
In our experiments, we employ FLEURS \cite{dataset_fleurs} and Dhwani \cite{dataset_in_w2v} datasets for spoken language identification. 
Additionally, we utilize Multilingual Librispeech dataset \cite{dataset_mls} for Automatic Speech Recognition (ASR). 
The FLEURS dataset consists of speech data for $102$ languages, with approximately $12$ hours of speech per language, derived from translated versions of $2009$ English Wikipedia sentences. All the translations are human generated with training, development and test containing $1500$, $150$ and $359$ sentences respectively. Each sentence was spoken by at least $3$ native speakers of the language.  

The Dhwani dataset encompasses multilingual speech data from $40$ Indian languages, downloaded from YouTube and the news platform \texttt{newsonair}. For our experiments, we use only the publicly accessible YouTube split, which consists of $12.6$k hours of speech in $22$ Indian languages. Unlike the FLEURS dataset, the Dhwani dataset is highly noisy and also contains substantial amounts of code-mixing, which challenges the label information based learning in the proposed \method{} method.

\subsection{Baseline systems}
\label{sec:baselines}
We compare \method{} framework with several other established benchmarks namely, (i) \textbf{wav2vec-2.0 (w2v)} model \cite{wave2vec2} pre-trained with SSL contrastive loss, (ii) \textbf{w2v-BERT} \cite{w2v_bert} model,   trained using the SSL MLM loss,  and (iii) \textbf{BEST-RQ }\cite{best_rq} model, which uses a random quantizer with the MLM loss.
Since the LASR approach is agnostic to the choice of the SSL objective function, we explore the combination of wav2vec-2.0, w2v-BERT and BEST-RQ model with the hard-triplet based LASR objective. All models are fine-tuned on the respective training split of the downstream task before evaluation. \\

\noindent \textbf{Implementation details:}
Most of the hyper-parameters are directly adopted from prior works \cite{w2v_bert,best_rq}. All the SSL baseline systems are pre-trained for $1.5$M epochs. For LASR training, the pre-trained SSL model at $1$M epochs is used as initialization, followed by $0.5$M steps of training with the LASR objective. All the models are  fine-tuned on the supervised training data for an additional $50$k epochs, with a batch size of $64$. We choose $\lambda$ from $\{2, 4, 8, 16\}$. The Adam optimizer \cite{adam_optimizer} is used in conjunction with a Transformer learning rate scheduler \cite{bert} that has $40$k warm-up steps. The learning rate is increased to $6e^{-4}$, followed by an inverse square root decay. We report mean of three runs for all the results.
\section{Results}
\label{sec:results}

\begin{table}[t!]
	\centering
	\small
	\begin{tabular}{lcccccc}
	    \toprule
	    \multicolumn{1}{l}{\textbf{Initialization}} & \multicolumn{3}{c}{\textbf{FLEURS}} & \multicolumn{3}{c}{\textbf{Dhwani}}  \\
	    \cmidrule(r){2-4} \cmidrule(r){5-7}
	    & Acc & F1 & EER & Acc & F1 & EER  \\
		\midrule
		Random & 52.8 & 46.5 & 2.1 & 55.3 & 25.1 & 14.9 \\
		BEST-RQ & 81.4 & 73.4 & 0.9 & 61.9 & 30.0 & 17.9 \\
		 + LASR   & 83.8 & 73.4 & 0.9 & 62.2  & 34.3 & 18.0    \\
		\bottomrule
	\end{tabular}
	\caption{\label{tbl:langid_supervised} Language recognition performance in supervised case. Fully supervised models have higher macro-F1 and EER.}
	\vspace{-0.25in}
\end{table}

The language recognition performance is measured using accuracy, equal error rate (EER) and macro-F1 score. These results are reported in  Table~\ref{tbl:langid_main}.
The languages in the test set (FLEURS/Dhwani) are split into two categories - a) the set of languages which overlap with the ones in the pre-training (denoted as $O$, $48$ classes in the FLEURS dataset and $5$ classes in the Dhwani dataset), and b) the set of languages which do not have any overlap with the set of languages in the pre-training data (denoted as $NO$, $54$ classes in the FLEURS dataset and $17$ classes in the Dhwani dataset). Further, the overall results   are also reported. The following are the key takeaways from the results reported in Table \ref{tbl:langid_main}.
\begin{itemize}
    \item On the FLEURS dataset, the LASR approach improves the BEST-RQ model relatively by $7.7$\%, $8.7$\%, and $58.3$\% in terms of accuracy, F1 and EER metrics, respectively. Similarly, on Dhwani dataset, the relative improvements from \method{} for BEST-RQ are $6.4$\%, $8.5$\%, and $5.0$\% on the above metrics. This trend is consistent across other pre-training methods as well. Thus, LASR framework improves over the baseline SSL results for both the datasets and for all the pre-training models (wav2vec-2.0, w2v-BERT and BEST-RQ).
    \item The improvements observed for the LASR approach are also consistent with  the overlap and the non-overlap subsets of the test data, and on all the three metrics reported.  
    \item For all the systems compared, the performance on the overlap set is consistently better than the non-overlap set. This indicates that, even when the pre-training objective did not explicitly use language labels (baseline SSL approaches), the language information is implicitly captured by the models. 
   
\end{itemize}

\noindent \textbf{Fully-Supervised Setting:} In Table \ref{tbl:langid_supervised}, we report the performance for  the scenario where a supervised pre-training is performed using the combined data of all the languages (pre-training and training data) with the cross-entropy loss. The label set is the union of the languages in the pre-training data and the fine-tuning data. The supervised model architecture is identical to the SSL and LASR models reported in Table \ref{tbl:langid_main}. We also experiment with three different initialization  choices for this supervised model - i) random initialization, ii) BEST-RQ model trained with SSL, and iii) BEST-RQ model trained with LASR objective.  
Our findings show that, on both the datasets, the fully-supervised setting does not achieve satisfactory results without weight initialization using a pre-trained model. While the accuracy of the supervised model improves over the SSL and LASR models in Table~\ref{tbl:langid_main}, EER and F1 scores are substantially worse for the supervised models. Nevertheless, the performance in this setting also improves with \method{} initialization.



\section {Discussion}
\label{sec:discussion}

\begin{table}[t!]
	\centering
	\begin{tabular}{lc}
	    \toprule
	    \textbf{Method} & \textbf{Accuracy} \\
		\midrule
		MLM  & 75.8 \\
        Hard-Triplet & 74.3 \\
        MLM  +  Triplet (Eq.~\ref{eq:triplet}) & 75.1 \\
        MLM  + GE2E (Contrastive) (Eq.~\ref{eq:ge2e}) & 76.2 \\
        MLM  + Hard-Triplet (Eq.~\ref{eq:hard}) & 80.4 \\
		\bottomrule
	\end{tabular}
	\caption{\label{tbl:langid_loss} Language identification accuracy with various loss functions. Hard-triplet loss performs best among all.}
	\vspace{-0.1in}
\end{table}

\begin{figure}[t!]
	\centering
	\includegraphics[width=0.95\linewidth]{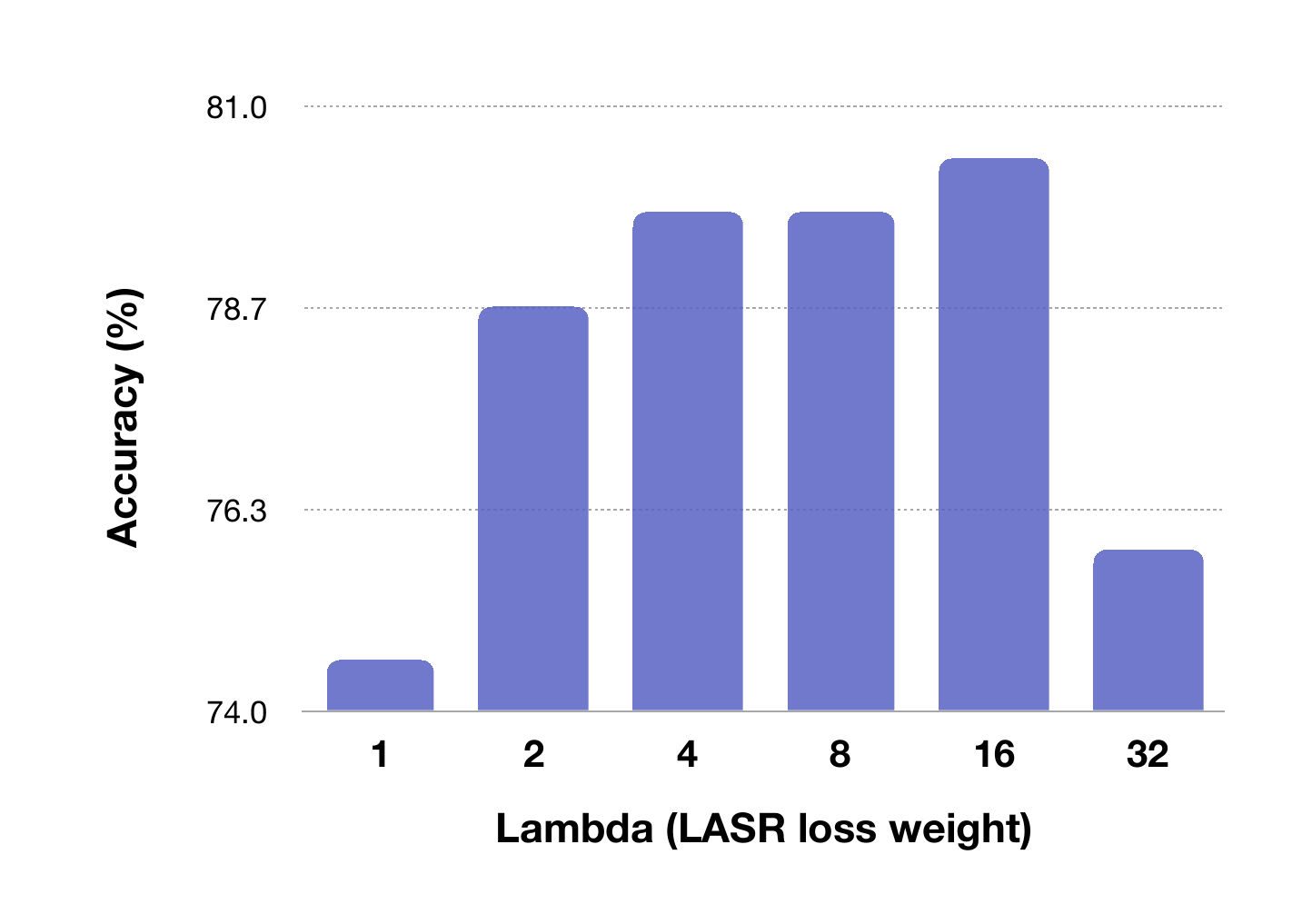}
	\vspace{-0.2in}
	\caption{\label{fig:lambda_plot}\small Accuracy for different choice of $\lambda$ in LASR objective (Eq.~\ref{eq:hard-triplet}). Slightly higher value of $\lambda (=16)$ is beneficial.}
	\vspace{-0.2in}
\end{figure}

\textbf{Effect of different optimization objectives} - Table \ref{tbl:langid_loss} compares various supervised loss functions. These experiments used the BEST-RQ \cite{best_rq} model evaluated on the FLEURS dataset. The first two experiments of Table~\ref{tbl:langid_loss} use only the MLM loss (SSL loss)  or only the supervised loss (Hard-triplet loss). The remaining experiments use the combined LASR loss (Eq.~\ref{eq:hard-triplet}). As seen here, the hard-triplet loss improves over other choices of semi-hard triplet loss or GE2E loss.  

\noindent  \textbf{Choice of supervised loss weight $\bm{\lambda}$} - 
For the hard-triplet loss in the LASR objective function (Eq.~\ref{eq:hard-triplet}), we have experimented with different choices of $\lambda$. These results are reported in Fig.~\ref{fig:lambda_plot}. The optimal choice of $\lambda$ is found to be $16$, which indicates that a higher weight for the supervised component is beneficial for the language recognition performance. However,  a larger  value, for example, $\lambda = 32$, degrades the performance. 

\begin{figure}[t!]
	\centering
	\includegraphics[width=\linewidth]{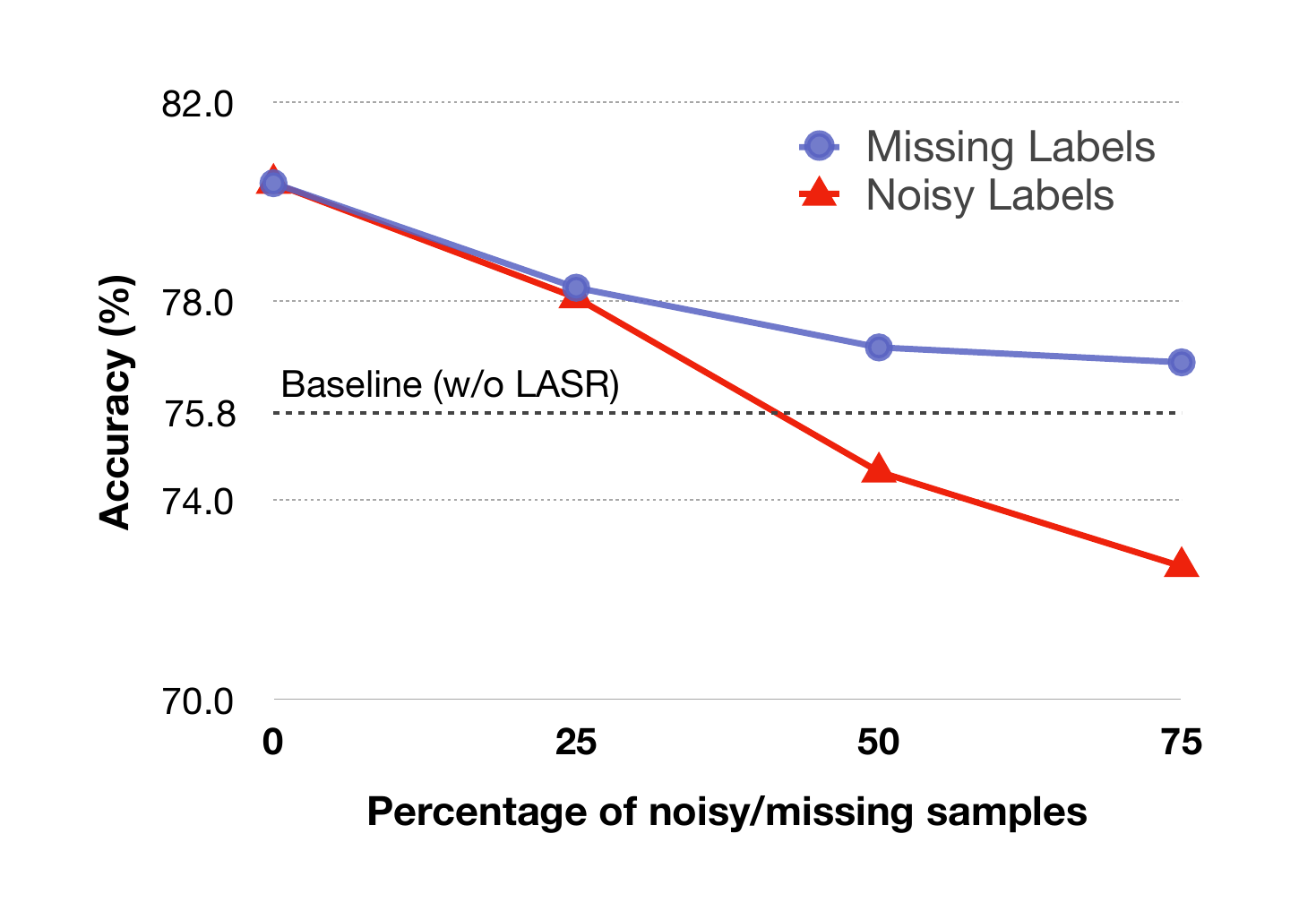}
	\vspace{-0.3in}
	\caption{\label{fig:noise_plot}\small Pre-training with missing or noisy labels. LASR is robust to missing and noisy language information in the data.}
	\vspace{-0.2in}
\end{figure}

 \noindent \textbf{Pre-training with missing/noisy labels} - All experiments reported thus far  used the language label information for the entire pre-training data.  We experiment with the robustness of the LASR approach for cases where the label information is either missing or noisy. For these experiments reported in Fig.~\ref{fig:noise_plot}, we assume $p$\% of the pre-training data to either have missing labels or have noisy labels (randomly corrupted to other language labels in pre-training set).  As expected, the language recognition performance degrades as $p$ increases. However, even when $75$\% of the pre-training data labels are missing,  LASR is significantly better than the baseline approach. The experiments highlight that the LASR approach can also yield performance improvements on pre-training data with noisy/missing labels.

\noindent \textbf{Impact on downstream ASR tasks} - In this section, we fine-tune LASR models on a semantic task, namely ASR. In particular, we run experiments for multilingual ASR on the MLS dataset. We follow a similar setup for ASR fine-tuning as was done in  \cite{best_rq}. To be more specific, we use the RNN-transducer model \cite{rnnt-graves}, where the decoder uses unidirectional LSTM. We do not employ shallow fusion with an external language model.
The ASR WER (\%) results are reported in Table \ref{tbl:asr_main}.   As seen in this table, the LASR based objective does not degrade the overall ASR performance even when the label information used in the LASR loss is an utterance-level non-semantic label. Thus, the representations learned using the LASR approach improve the language recognition tasks without any degradation on semantic tasks such as ASR.

\begin{table}[t!]
	\centering
	\small
	\begin{tabular}{lccccccccc}
	    \toprule
	    \multicolumn{1}{c}{\textbf{Method}} & \multicolumn{6}{c}{\textbf{Languages}} & \textbf{Avg} \\
        \cmidrule(r){2-7}
         & de & en & es & fr & it & nl  \\
		\midrule
        w2v-BERT  & 4.0 & 6.2 & 4.0 & 4.7 & 8.9 & 10.6  & 7.2\\
         + \method{} & 4.0 & 6.2 & 4.8 & 4.8 & 8.9 & 10.0  & 7.2\\
        \midrule
        BEST-RQ  & 3.9 & 6.2 & 3.8 & 4.8 & 8.8 & 9.3  & 7.0 \\
		+ \method{} & 4.1 & 6.2 & 4.3 & 4.8 & 9.0 & 9.6  & 7.1  \\
		
		\bottomrule
	\end{tabular}
	\caption{\label{tbl:asr_main} WER (\%) for ASR on Multilingual LibriSpeech. Adding the non-semantic LASR objective during pre-training does not degrade performance on semantic tasks such as ASR.}
	\vspace{-0.25in}
\end{table}

\section{Conclusion}
\label{sec:conclusion}
In this paper, we introduce a method for enhancing self-supervised speech representation learning by incorporating non-semantic language label information. Our proposed approach, Label Aware Speech Representation (LASR) learning, utilizes a triplet-based objective in addition to the self-supervised loss function. The results from language recognition experiments demonstrate that the LASR approach provides substantial overall improvements, particularly on subsets of test data that do not overlap with pre-training languages. Additionally, experiments on the automatic speech recognition (ASR) task indicate that the LASR model produces speech representations that do not compromise performance for semantic tasks.




\bibliographystyle{IEEEtran}
\bibliography{mybib}

\end{document}